# TS-MAMP: A Remanufactured Agricultural Robot with Second-Life EV Components and NMS-Free On-Device Weed Detection

Weijie Shi[1,†], Zicheng Xu[2,3,*,†], Zhenbang Cheng[1,*], Haoran Xuan[1], Mingbo Duan[1], Gan Ge[1]

[1] *School of Mechanical and Automotive Engineering, West Anhui University, Lu'an, 237000, China*

[2] *School of Automation, Nanjing University of Information Science and Technology, Nanjing, 210000, China*

[3] *School of Engineering, Westlake University, Hangzhou, 310030, China*

[†] These authors contributed equally to this work.

[*] Correspondence: 202542000226@nuist.edu.cn (Z.X.); chzhbang@qq.com (Z.C.).

*Abstract*—**Agriculture 4.0 robotic systems improve field-operation efficiency yet remain too capital-intensive for many fragmented smallholder farms. Meanwhile, a growing number of retired low-speed electric-vehicle (LSEV) powertrains retain functional electromechanical value but are often destructively recycled. This paper presents TS-MAMP (Telescopic-Sleeve Modular Agricultural Mobile Platform), a remanufactured robot built under 3R (reduce, reuse, recycle) circular-economy principles. Retired 48 V brushless-DC (BLDC) hub motors are paired via back-electromotive-force (back-EMF) matching, and lead-acid battery modules retaining 60%–80% of their rated capacity are integrated with an active equalizer to mitigate inter-module voltage imbalance. Together, these reused components reduce the powertrain-and-chassis bill-of-materials (BOM) cost by approximately 60%, to below USD 450 (perception and weeding modules excluded). The truss chassis provides ≥200 kg static load, continuously adjustable track width from 1200 mm to 2000 mm, and ≤5-minute module changeover. For perception, a non-maximum suppression (NMS)-free YOLOv10n detector optimized with data augmentation, spatial and channel attention, and negative-sample training achieves 82.43% mean average precision (mAP)@0.5 and 59.61% mAP@0.5:0.95 on the independent test set. The detector is deployed via FP16 TensorRT on an NVIDIA Jetson Nano, demonstrating the feasibility of on-device inference. TS-MAMP demonstrates that retired EV components, under modest screening, can be re-engineered into affordable, AI-enabled agricultural robots—opening a remanufacturing pathway for the smallholder fields that commercial automation leaves unserved.**

## I. Introduction

Many high-throughput Agriculture 4.0 robotic systems are designed for large, standardized farmland and require substantial capital investment. The NEXAT gantry, with its ~1,100 hp drivetrain and 200 t/h harvesting modules, represents a multimillion-pound investment designed for very large-scale commercial farms [1]–[3]—illustrating the stark gap between high-end automation and the small, fragmented holdings typical of smallholder agriculture [4]. Modular robots such as Thorvald II have improved configurability across field, greenhouse, and specialty-crop environments [5], [6], while commercial systems like Naïo Technologies' Ted have demonstrated the progress of autonomous mechanical weeding in specialty crops [7]. Yet smallholder adoption remains constrained by capital cost, financial risk, limited credit access, and uncertainty about returns [8]–[10].

Meanwhile, rapid transport electrification is generating a growing stream of retired EV components. In China, retired NEV power batteries are projected to exceed one million tonnes annually by 2030 [11], and comprehensive reviews have confirmed the viability of repurposing these retired batteries for second-life applications [12]. Additionally, electric traction motors contain valuable assemblies that are often lost when end-of-life treatment relies on destructive recycling [13]–[15]. Second-life studies further indicate that retired batteries retaining substantial residual capacity can remain suitable for lower-demand applications [16]–[19], suggesting an opportunity to connect reuse-oriented remanufacturing with low-speed agricultural robotic platforms.

In this work, we present TS-MAMP, a remanufactured agricultural robot built from second-life LSEV components under 3R circular-economy principles [20]–[23]. Unlike high-end gantry or modular platforms that require new components and substantial capital, TS-MAMP targets fragmented smallholder terrain through a truss-style, dimensionally reconfigurable chassis powered by retired BLDC motors and lead-acid battery modules, paired with an NMS-free YOLOv10n perception system [24] that removes the need for a separate NMS post-processing step on edge hardware. The contributions are threefold: (1) a second-life EV powertrain screening and modular chassis integration approach that reduces the powertrain-and-chassis BOM cost by approximately 60% to below USD 450 (perception and weeding modules excluded); (2) prototype-level mechanical verification of ≥200 kg static load, continuously adjustable track width from 1200 mm to 2000 mm, and ≤5 min tool-module changeover; and (3) an NMS-free YOLOv10n pipeline achieving 82.43% mAP@0.5 and 59.61% mAP@0.5:0.95 on the independent test set, with FP16 TensorRT deployment on a Jetson Nano for on-device inference.

## II. System Design

### A. Green Remanufacturing and EV Component Qualification

The TS-MAMP platform was designed and fabricated following 3R circular-economy principles [20]–[23]. The remanufacturing workflow prioritized direct reuse of functional electromechanical assemblies salvaged from retired low-speed electric vehicles (LSEVs), minimizing raw-material extraction and manufacturing energy. Paired DJZ48-10C1 48 V/350 W BLDC hub motors were selected as the traction actuators. Salvaged units were matched by back-EMF waveform and internal resistance prior to installation, to reduce left–right drive asymmetry between the left and right drive units.

For energy storage, retired 12 V graphene-modified lead-acid battery modules were collected from LSEV scrapping stations and evaluated with reference to the capacity and safety test methods specified in GB/T 32620.1-2016 [25]. Modules retaining 60%–80% of rated capacity were selected and reassembled into a 48 V (four-series) / 20 Ah traction pack. The 60%–80% capacity-retention range was determined from preliminary discharge characterization of 20 retired modules and was used as a practical screening range to ensure sufficient residual capacity for intermittent, relatively low-current agricultural operation. This approach redirects retired battery modules from uncontrolled disposal into a controlled reuse pathway, reducing the risk of electrolyte leakage and heavy-metal contamination associated with improper end-of-life treatment. The qualified powertrain configuration is summarized in Table I.

TABLE I. Configuration of Remanufactured Powertrain and Chassis

| Component | Source (Retired EV) | Specification in Robot |
|---|---|---|
| Energy Storage | 12 V 20 Ah Graphene-modified Lead-acid Battery × 4 | 48 V System (4 Series) |
| Traction Motor | 48 V 350 W Hub Brake Motor | 48 V 350 W BLDC Actuator |
| Transmission | 38-Link Chain & 48T/16T Gear | Chain Drive (Reduction Ratio 3:1) |
| Control Unit | Brushless EV Controller | Integrated Dual-Mode PWM Driver |
| Chassis Node | Front Wheel Hub (Model 21-44-20) | Differential Steering System |

The fully assembled prototype was built using retired components, an in-house fabricated frame, and off-the-shelf control electronics, with a prototype-level direct hardware BOM cost below USD 450 covering the traction powertrain and chassis only (motors, batteries, frame, controller, battery equalizer, and mechanical drivetrain; the perception stack—Jetson Nano and IMX219 camera—and the laser-weeding module excluded), approximately 60% lower than an equivalent new-component baseline comprising commercially sourced 48 V BLDC motors, new lead-acid traction batteries, and a matching frame structure.

### B. Mechanical Architecture and Drivetrain

The mechanical platform was designed as a twin-parallel bracket truss frame joined by a central support base. The load-bearing structure combines a high-strength aluminum-alloy portal frame with Q235 low-carbon-steel beams, providing a compromise between structural rigidity, reduced frame weight, fabricability, and material cost. The rated static load capacity of ≥200 kg was verified through physical prototype loading. The powertrain adopts a chain transmission rather than a belt drive to avoid belt slippage under dusty and humid field conditions. A 3:1 sprocket reduction (16-tooth driving sprocket, 48-tooth driven sprocket) nominally provides threefold torque multiplication at the wheel hubs, creating a torque safety margin that compensates for the expected electromechanical attenuation of second-life motors. This mechanical amplification allows the aged BLDC actuators to operate below their peak torque demand during normal traction. The overall mechanical architecture and powertrain layout are illustrated in Fig. 1.

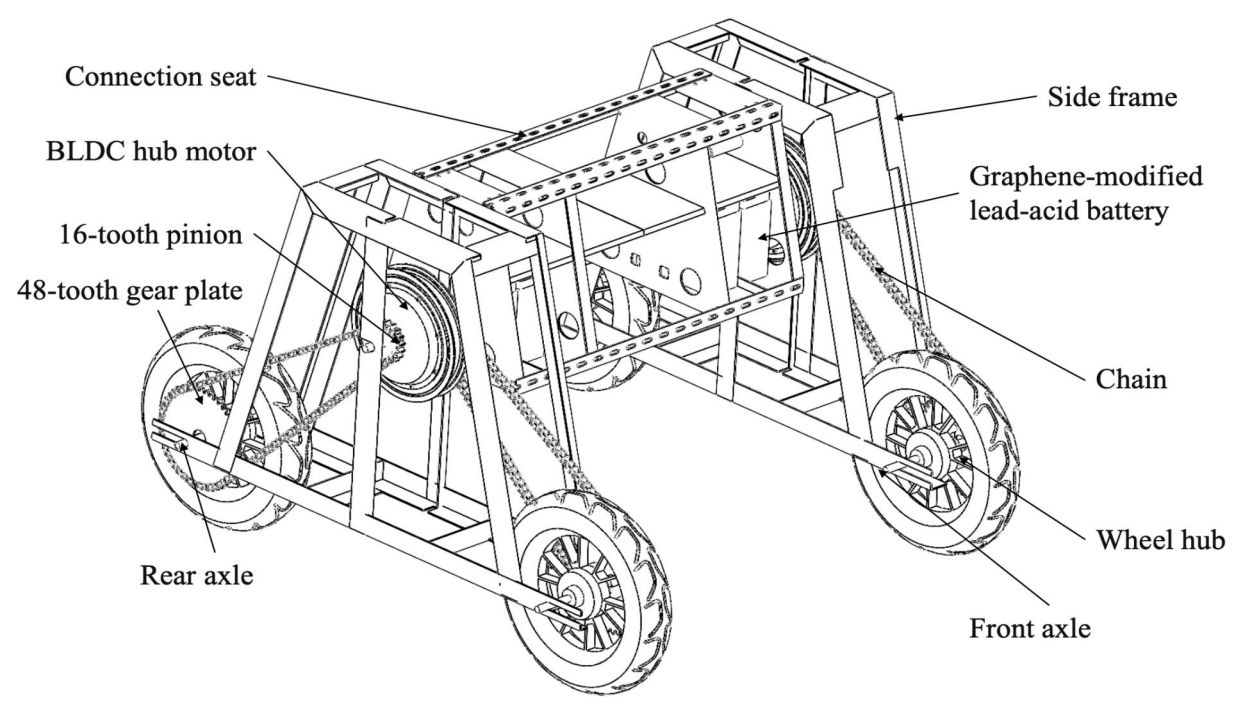

Figure 1. Mechanical architecture and powertrain components.

To accommodate variable crop row spacing and module reconfiguration, the frame integrates two independent adjustment mechanisms: (i) a continuous telescopic-sleeve system enabling continuously adjustable track width from 1200 mm to 2000 mm, and (ii) an independent sliding-groove system providing lateral module-width adjustment from 300 mm to 600 mm at ±5 mm positioning accuracy. Both mechanisms are secured by vibration-resistant locking bolts, allowing complete tool-module changeover within 5 minutes.

### C. Multifunctional Modules and Laser Weeding

The TS-MAMP platform utilizes standardized interfaces and rapid connection mechanisms to enable configuration of multifunctional modules. A primary configuration is the laser weeding sub-module, a non-chemical alternative to herbicide-based weeding [26], [27]. This module employs a 3-DOF Delta parallel manipulator comprising three identical limbs actuated by 42 mm hybrid stepper motors with 0.48 N·m holding torque. A parallelogram guide linkage provides positioning of a four-stage laser system with 4 mm spot diameter and 50 W optical output power. The laser assembly includes a protective enclosure designed with reference to the protective housing requirements of GB/T 7247.1-2024 [28].

Laser activation is precisely controlled via the Modbus-RTU communication protocol and integrated with safety interlock devices. Laser emission is inhibited when the platform speed remains below 0.1 m/s for longer than a predefined dwell-time threshold or when no valid target is detected, thereby reducing the risk of excessive localized heating during prolonged dwell and unintended laser exposure. The multifunctional module configuration is shown in Fig. 2.

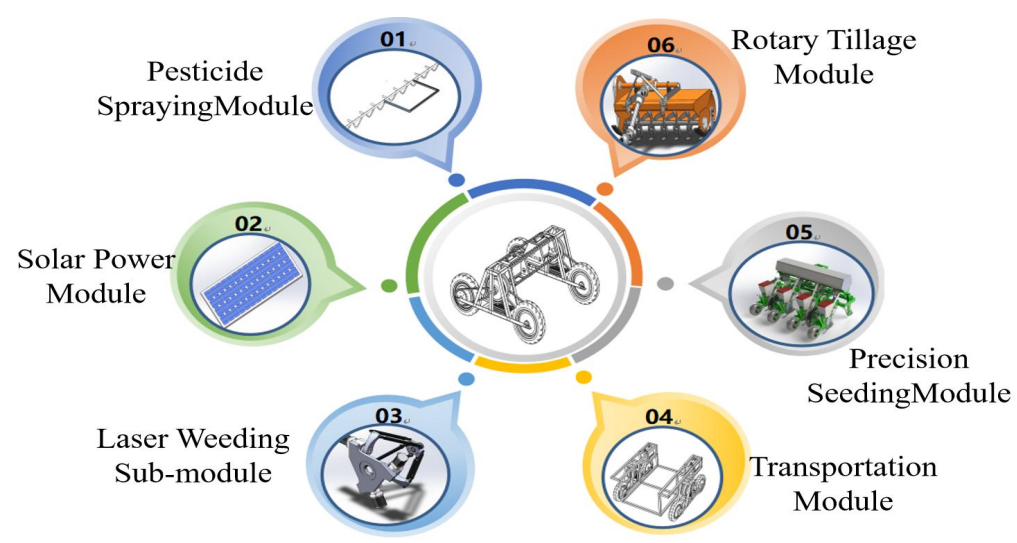

(a) Module schematic

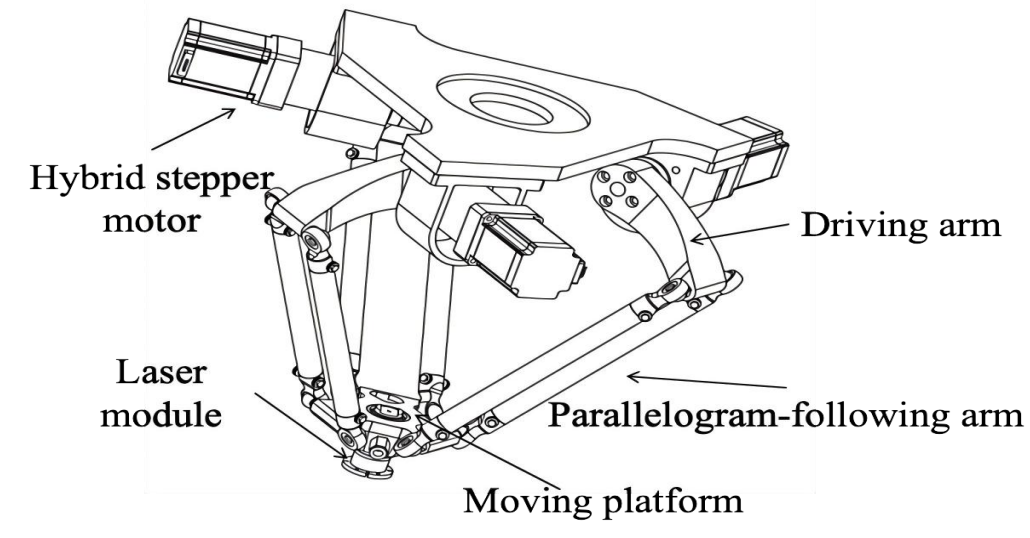

(b) Laser weeding submodule structure

Figure 2. Multifunctional module configuration.

### D. Electronic Control and Drive System

The electronic control system integrates power management, drive control, and onboard perception, enabling cost-effective repurposing of retired EV components for operator-supervised field operation. The powertrain uses salvaged DJZ48-10C1 BLDC hub motors (48 V, 350 W), which are equipped with integrated mechanical expansion brakes, as illustrated in Fig. 3.

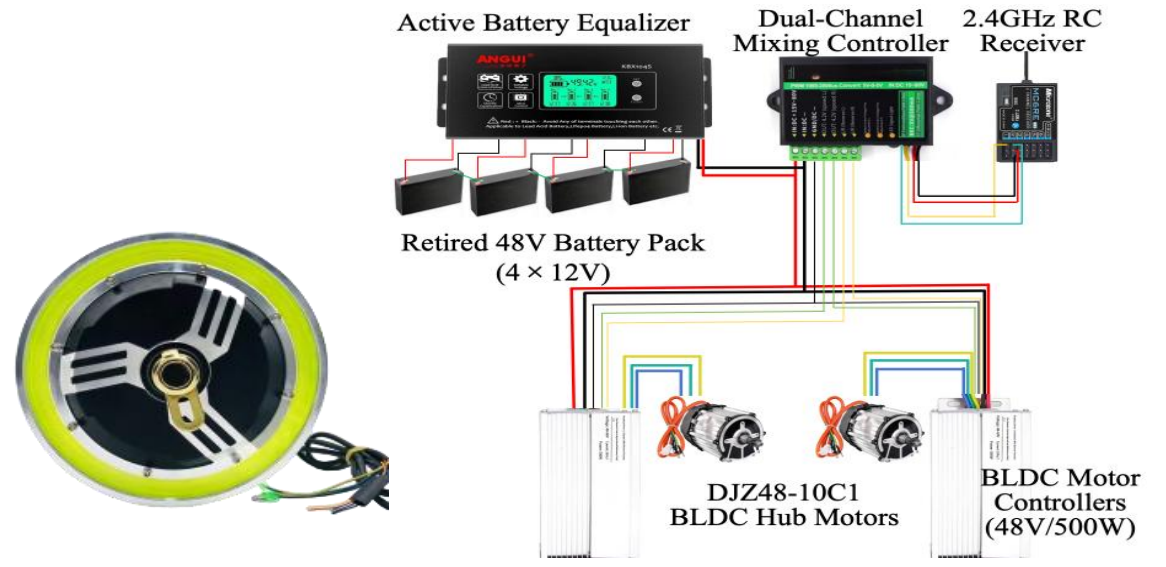

(a) DJZ48-10C1 BLDC hub brake motor (b) Workflow schematic

Figure 3. Electronic control and drive module.

A dual-channel mixing controller serves as the central electronic interface between the wireless command system and the drive actuators. It translates PWM command signals from a 2.4 GHz Microzone MC6RE radio receiver into calibrated analog set-points for two 500 W BLDC motor speed controllers while maintaining a low-latency manual override via the wireless link for operational safety during system debugging and field maneuvers.

To ensure operational safety and alleviate voltage imbalance in the retired batteries, the system is equipped with an ANGUI KBX104S active battery equalizer. According to the manufacturer's technical support, the balancing activation threshold is an inter-module voltage difference of 100 mV. For thermal management, modules are physically separated by 15 mm gaps within a ventilated chassis to facilitate passive air-cooling dissipation.

## III. Dataset and Experimental Evaluation

### A. *Wanxi Crop-Weed Dataset*

To address the domain gap between public crop-weed datasets and local field conditions, we constructed the Wanxi Crop-Weed Dataset in Lu'an, Anhui Province. Images were acquired using the platform's onboard Sony IMX219 CMOS camera at 1920 × 1080 resolution under natural illumination of 20,000–100,000 lux. All data were captured during the 3–5 leaf seedling stage of pak choi, a visually challenging stage because of substantial morphological similarity between crops and weeds. The 675 images were split before augmentation into 439 training, 169 validation, and 67 test images (65%/25%/10%). Augmentation was applied only to the training set; validation was used for ablation and model selection, and the held-out test set for final evaluation. Annotation followed a plant-wise bounding-box protocol covering five classes: pak choi *(Brassica rapa* subsp. *chinensis*) and four associated weeds: *Echinochloa crus-galli*, *Eleusine indica*, *Digitaria sanguinalis*, and *Portulaca oleracea*.

### B. *NMS-Free Perception Pipeline and Optimization*

We adopted YOLOv10n with consistent dual assignments for NMS-free training [24]. An auxiliary one-to-many head provides rich supervisory signals during training, while a primary one-to-one head assigns a single prediction to each ground-truth object. During inference, only the optimized one-to-one head is retained, eliminating the need for separate NMS post-processing and thereby simplifying the inference pipeline for edge deployment. Training was performed on an NVIDIA RTX 3090 (24 GB) workstation; inference was deployed on an NVIDIA Jetson Nano using TensorRT with FP16 enabled, consistent with prior TensorRT-based optimization studies on the Jetson Nano [29].

To address slow convergence and false-positive detections in complex seedling-stage scenes, a two-stage optimization strategy was applied. In Stage I, only the 439-image training partition was augmented offline using mosaic composition and photometric and geometric transformations, yielding 6,020 training samples. In Stage II, a spatial attention module (SAM) and a channel attention module (CAM) were progressively incorporated into the detector, followed by the addition of 500 target-free background images. This increased the final training set to 6,520 samples, of which 7.67% were negative background images. No augmented or negative samples were introduced into the validation or test partitions.

Table II reports the cumulative ablation results evaluated on the same validation set for all configurations. The NMS-free YOLOv10n baseline achieved 62.15% mAP@0.5 and 45.32% mAP@0.5:0.95.

TABLE II. CUMULATIVE ABLATION RESULTS ON THE VALIDATION SET

| Configuration | Train samples | Aug. | SAM | CAM | Neg. | P (%) | R (%) | mAP50 (%) | mAP50-95 (%) | Δ mAP50 (pp) | Δ mAP50-95 (pp) |
|---|---|---|---|---|---|---|---|---|---|---|---|
| Baseline | 439 | — | — | — | — | 72.28 | 67.34 | 62.15 | 45.32 | — | — |
| + Data Aug. | 6,020 | ✓ | — | — | — | 77.03 | 71.29 | 69.46 | 49.37 | +7.31 | +4.05 |
| + SAM | 6,020 | ✓ | ✓ | — | — | 79.65 | 74.13 | 74.22 | 52.81 | +4.76 | +3.44 |
| + CAM | 6,020 | ✓ | ✓ | ✓ | — | 81.38 | 75.61 | 76.64 | 55.06 | +2.42 | +2.25 |
| + Negative Samples | 6,520 | ✓ | ✓ | ✓ | ✓ | 83.31 | 78.41 | 80.87 | 58.41 | +4.23 | +3.35 |

Note: All configurations were trained for 100 epochs, and the reported metrics correspond to epoch 100. Δ values are step-wise improvements relative to the preceding row.

Figure 4 shows the validation mAP@0.5 and mAP@0.5:0.95 trajectories for the baseline and four

successive optimization configurations over 100 epochs.

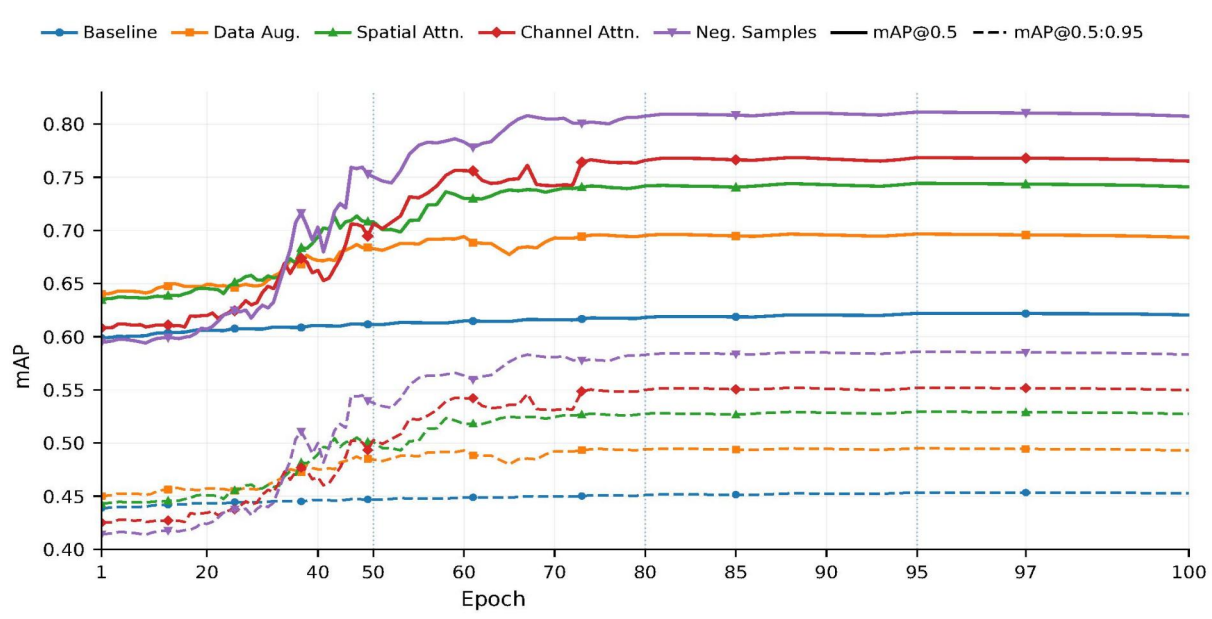

Figure 4. Validation mAP curves for the baseline and four successive optimization configurations over 100 epochs using nonuniform epoch spacing.

Data augmentation increased mAP@0.5 from 62.15% to 69.46% (+7.31 pp). Subsequent incorporation of SAM, CAM, and negative-sample training further increased mAP@0.5 to 74.22%, 76.64%, and 80.87%, respectively, yielding a cumulative improvement of 18.72 percentage points over the baseline. The corresponding mAP@0.5:0.95 increased from 45.32% to 58.41% (+13.09 pp). After selecting the final configuration based on validation performance, the resulting detector was evaluated on the held-out 67-image test set, achieving 84.37% precision, 79.27% recall, 82.43% mAP@0.5, and 59.61% mAP@0.5:0.95.

Loss curves of both the auxiliary and primary heads showed steady downward trends and rapid stabilization across the evaluated configurations (Fig. 5). Field detection results from the deployed Jetson Nano platform are shown in Fig. 6.

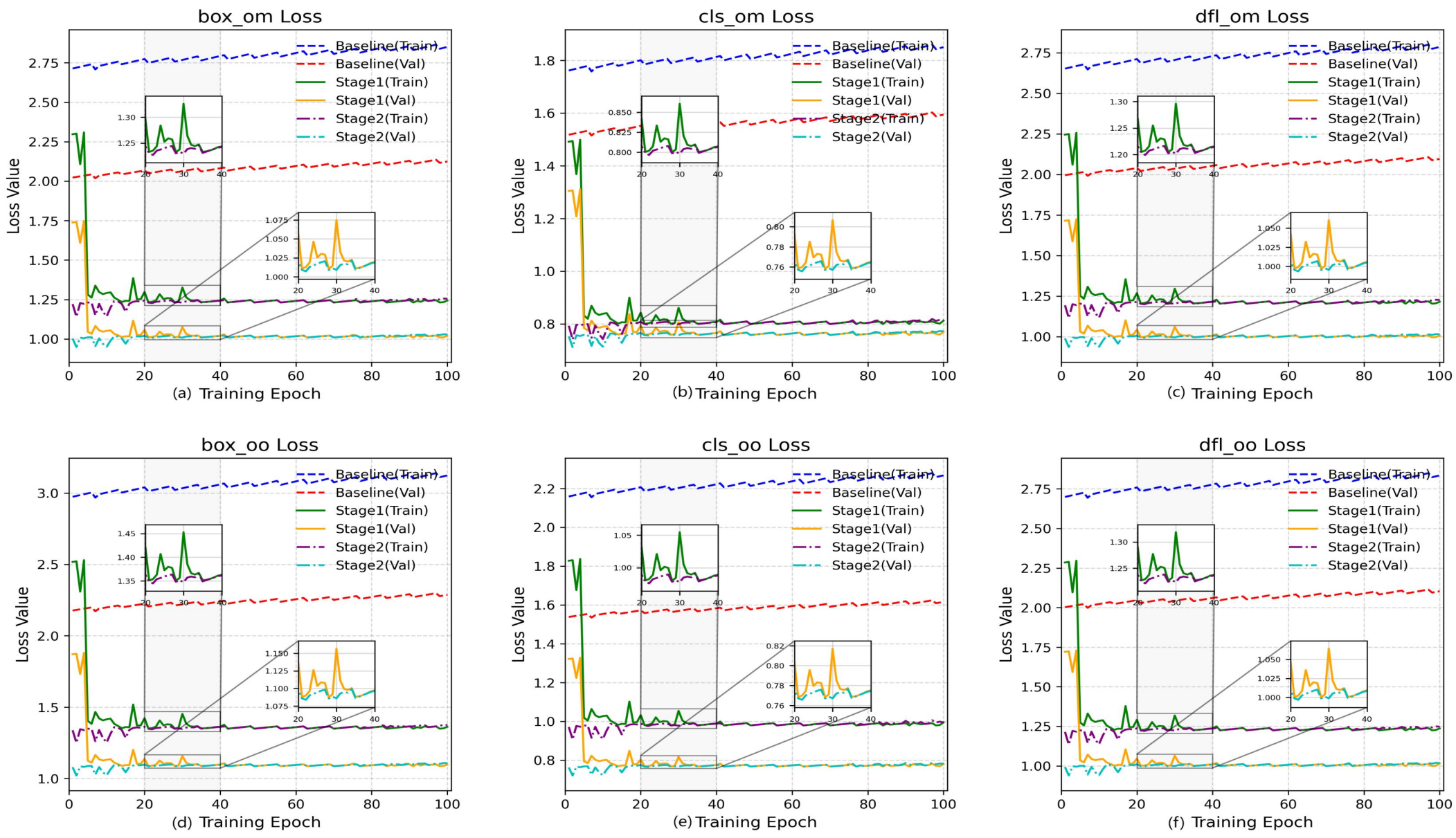

Figure 5. Training and validation loss comparison across the baseline and the two optimization stages.

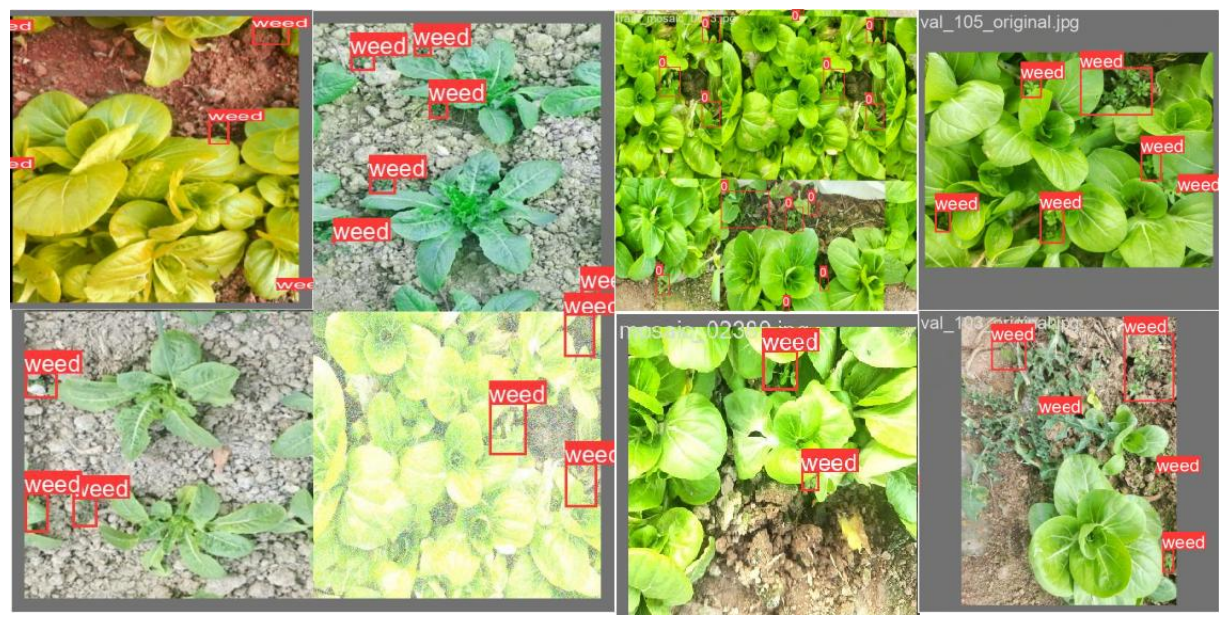

Figure 6. YOLOv10n weed detection results in field conditions.

## C. Field Validation

A physical prototype of the TS-MAMP was constructed and evaluated in field trials in Lu'an, Anhui (Fig. 7). The tests demonstrated the structural integrity of the modular truss chassis and the operational feasibility of the repurposed (second-life) powertrain under the tested field conditions. The standardized interfaces facilitated rapid switching between functional modules. Integrated with the optimized YOLOv10n perception system, the platform demonstrated field mobility and on-device inference capability, supporting the feasibility of repurposing retired EV components for low-cost agricultural automation.

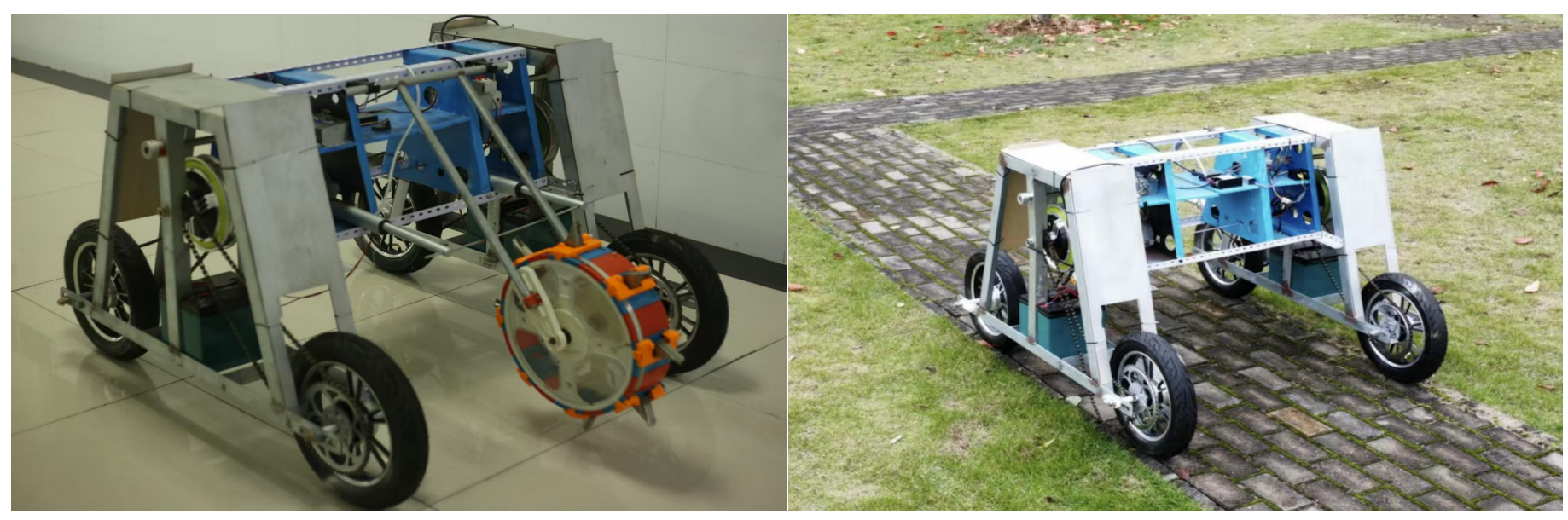

(a) Physical prototype (b) Field test

Figure 7. Field validation of the TS-MAMP prototype.

## IV. Conclusion

This work demonstrates that repurposed retired EV powertrains can serve as a viable foundation for affordable, intelligent agricultural machinery. By applying 3R principles and screening retired battery modules with reference to GB/T 32620.1-2016, we integrated repurposed 48 V BLDC motors and graphene-modified lead-acid batteries into a functional platform, reducing the powertrain-and-chassis BOM cost by approximately 60% to below USD 450 (perception and weeding modules excluded). On the perception side, the NMS-free YOLOv10n detector achieved 82.43% mAP@0.5 and 59.61% mAP@0.5:0.95 on the independent test set and was deployed on the resource-constrained Jetson Nano via FP16 TensorRT. These results establish TS-MAMP as a practical pathway linking the circular reuse and remanufacturing of retired EV components with precision agriculture.

The current approach operates within certain boundaries, particularly the capacity heterogeneity of retired battery modules and the absence of multi-modal sensing. Future work will prioritize integrating active balancing functionality into a comprehensive BMS to manage capacity variation and incorporating LiDAR–vision sensor fusion to improve perception under variable field and illumination conditions.

## Funding

This study was supported by the Anhui Provincial Key Research and Development Project (2024AH051996), the Anhui Provincial Teaching Reform Research Project (2023JYXM0695), and the Anhui AI+Education Curriculum Project (2024AIJY327).